\documentclass[letterpaper, 10 pt, conference]{ieeeconf}
\IEEEoverridecommandlockouts                              % This command is only needed if 
\usepackage{caption}
\usepackage{xcolor}
\usepackage{cite}
\usepackage{graphicx}
\usepackage{amsmath}
\usepackage{textcomp}
\usepackage{algorithm}
\usepackage{amssymb}
\usepackage{textcomp}
\usepackage{algpseudocode}
\usepackage{tabularx} % tables
\usepackage{mathrsfs} 
\usepackage{soul,xcolor}
\usepackage[normalem]{ulem}

\usepackage[switch]{lineno}

\makeatletter
\let\NAT@parse\undefined
\makeatother
\usepackage[hidelinks]{hyperref}
\hypersetup{
    colorlinks=true,
    linkcolor=blue,
    citecolor=blue,
    filecolor=magenta,      
    urlcolor=cyan,
    pdftitle={Overleaf Example},
    pdfpagemode=FullScreen,
    }

\title{\LARGE \bf 
Calibration-Free Gas Source Localization with Mobile Robots: Source Term Estimation Based on Concentration Measurement Ranking}
\author{Wanting Jin, Agatha Duranceau, İzzet Kağan Erünsal and Alcherio Martinoli% <-this % stops a space
\thanks{The authors are with the Distributed Intelligent Systems and Algorithms Laboratory, School of Architecture, Civil and Environmental Engineering, \'{E}cole Polytechnique F\'{e}d\'{e}rale de Lausanne (EPFL), 1015 Lausanne, Switzerland. This work is funded by the Swiss National Science
Foundation under grants 200020\_175809 and 10.001.747. Additional information about the
research can be found here: \url{https://disal.epfl.ch/research/gassensingstructure/}
}
}

\begin{document}
%\linenumbers
\setstcolor{red}
\maketitle
\thispagestyle{empty}
\pagestyle{empty}

%%%%%%%%%%%%%%%%%%%%%%%%%%%%%%%%%%%%%%%%%%%%%%%%%%%%%%%%%%%%%%%%%%%%%%%%%%%%%%%%
%%%%%%%%%%%%%%%%%%%%%%%%%%%%% ABSTRACT %%%%%%%%%%%%%%%%%%%%%%%%%%%%%%%%%%%%%%%%%
%%%%%%%%%%%%%%%%%%%%%%%%%%%%%%%%%%%%%%%%%%%%%%%%%%%%%%%%%%%%%%%%%%%%%%%%%%%%%%%%
\begin{abstract}
Efficient Gas Source Localization (GSL) in real-world settings is crucial, especially in emergency scenarios. Mobile robots equipped with low-cost, in-situ gas sensors offer a safer alternative to human inspection in hazardous environments. Probabilistic algorithms enhance GSL efficiency with scattered gas measurements by comparing gas concentration measurements gathered by robots to physical dispersion models. However, accurately deriving gas concentrations from data acquired with low-cost sensors is challenging due to the nonlinear sensor response, environmental dependencies (e.g., humidity, temperature, and other gas influences), and robot motion. Mitigating these disturbance factors requires frequent sensor calibration in controlled environments, which is often impractical for real-world deployments. To overcome these issues, we propose a novel feature extraction algorithm that leverages the relative ranking of gas measurements within the dynamically accumulated dataset. By comparing the rank differences between gathered and modeled values, we estimate the probabilistic distribution of source locations across the entire environment. We validate our approach in high-fidelity simulations and physical experiments, demonstrating consistent localization accuracy with uncalibrated gas sensors. Compared to existing methods, our technique eliminates the need for gas sensor calibration, making it well-suited for real-world applications.% By eliminating the need for frequent and accurate calibration, the proposed algorithm enhances robustness to sensor drift, nonlinearity, and environmental variations, making it well-suited for real-world gas source localization tasks.
\end{abstract}

%%%%%%%%%%%%%%%%%%%%%%%%%%%%%%%%%%%%%%%%%%%%%%%%%%%%%%%%%%%%%%%%%%%%%%%%%%%%%%%%INTRODUCTION%%%%%%%%%%%%%%%%%%%%%%%%%%%%%%%%%%%%%%%%%%
\section{Introduction}
Gas leak detection is critical for industrial safety, environmental protection, and public health~\cite{jing_recent_2021}. Localizing the source of the leak is often time-sensitive, especially in emergency scenarios. However, measuring gas concentrations in large areas is time-consuming, as in-situ gas sensors provide only point measurements at specific locations and times. Consequently, efficiently estimating the source with only a limited number of measurements is of paramount importance. Probabilistic algorithms facilitate Gas Source Localization (GSL) with scattered gas measurements by incorporating knowledge of gas plume models \cite{keats_bayesian_2007}. They iteratively update the belief of each candidate source position by assessing the alignment between the \textbf{measured gas concentrations} derived from the sensor readings and the \textbf{estimated gas concentrations} derived from gas plume models. However, real-world GSL is challenging because of the limited faithfulness of plume models and reliability of concentration measurements. In this work, we propose addressing the \textbf{model-to-measurement gap} with a gas feature that reduces dependency on precise plume models while effectively utilizing raw sensor readings to retain information for probabilistic GSL algorithms.
\begin{figure}
\vspace{0.2cm}
    \centering
    \includegraphics[width=\linewidth]{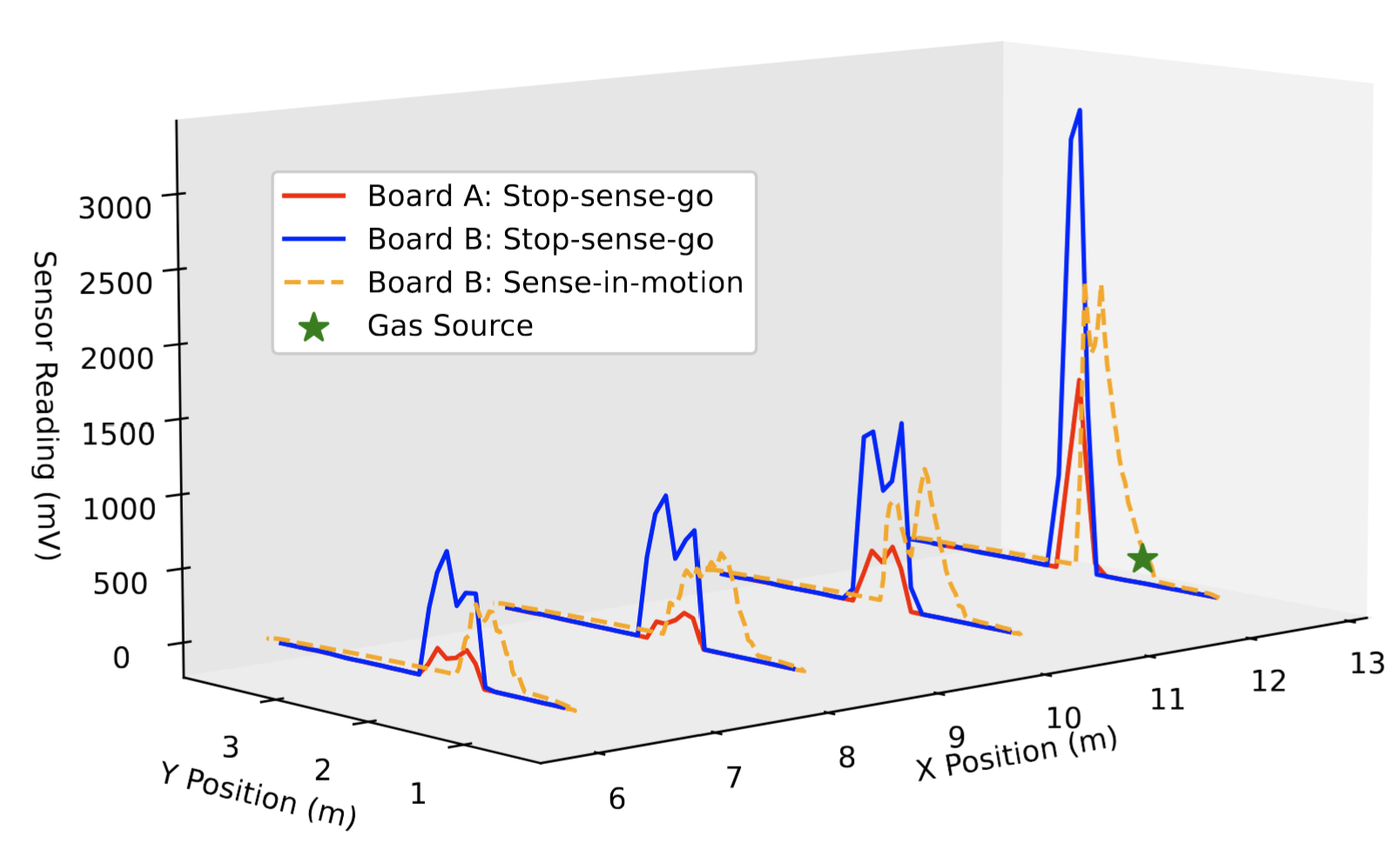}
    \caption{{Gas measurements collected using different sensor boards (\textit{A} and \textit{B}) and gas sampling strategies (\textit{stop-sense-go} and
    \textit{sense-in-motion}). The difference in sensor responses highlights the impact of both sensor properties and robot motion on gas measurement consistency.}}
    \label{fig:sensor_scan}
    \vspace{-0.6cm}
\end{figure}

%The critical real-world challenges in GSL are often overlooked by many existing works solely carried out in simulation~\cite{hutchinson_adaptive_2017,ruiz_gas_2024, nanavati_distributed_2024, an_receding-horizon_2022, shi_autonomous_2025}. In particular, many existing approaches rely on two unrealistic assumptions: (i) obstacle-free environments~\cite{hutchinson_adaptive_2017}\cite{ruiz_gas_2024}, and (ii) a simplified sensor model that directly converts sensor readings into accurate gas concentration values \cite{nanavati_distributed_2024, an_receding-horizon_2022, shi_autonomous_2025}. {\color{red} With assumption (i)}, the gas dispersion under different source locations is modeled analytically, assuming a steady regime. In realistic settings, however, obstacles in the environment can significantly influence the gas dispersion pattern, hindering analytical solutions.

Given the complexity of conducting GSL experiments in the real world, many existing works are performed solely in simulation~\cite{hutchinson_adaptive_2017,ruiz_gas_2024, nanavati_distributed_2024, an_receding-horizon_2022, shi_autonomous_2025}. However, two critical challenges are often overlooked, which limits the transfer of these methods to real-world applications: (i) complex environmental layouts with obstacles, and (ii) the slow, nonlinear response of physical gas sensors to changes in gas concentration. In~\cite{hutchinson_adaptive_2017}\cite{ruiz_gas_2024}, the environment is assumed to be obstacle-free, and therefore the gas dispersion can be modeled analytically under a steady regime assumption. In realistic settings, however, obstacles can significantly influence the gas dispersion pattern, making analytical solutions infeasible.
To address challenge (i), one promising approach proposed in our previous work is to use a Data-Driven Plume Model (DDPM) as a surrogate gas plume model for built environments~\cite{jin_towards_2023}. DDPM is a learning-based method that predicts the gas concentration distribution given a hypothesized source location and environmental layout.  
%^{\color{red} \st{However, the generalization capacity of this approach remains constrained by the diversity of the training data.}} 
Regarding challenge (ii), in \cite{nanavati_distributed_2024, an_receding-horizon_2022, shi_autonomous_2025} gas sensor outputs are treated as accurate concentration measurements perturbed only by white noise. In practice, however, physical gas sensors such as Metal OXide (MOX) sensors, depend on direct exposure and chemical interaction with the target gas, and therefore exhibit slow and nonlinear response to concentration changes.
 %{\color{red}\st{which is influenced by factors such as manufacturing variation, environmental conditions, and sensor aging}}~\cite{francis_gas_2022}.
 In addition, these sensors usually have slow response and recovery dynamics, which can lead to spatial misalignment when measurements are collected while the robot is moving.

To illustrate this, Fig.~\ref{fig:sensor_scan} highlights the impact of sensor properties and robot motion on gas measurements. The solid red and blue lines show outputs from two MiCS-5521 MOX sensor boards in a \textit{stop-sense-go} mode, where the robot pauses at each sampling point. The orange dashed line shows readings from the same sensor board as the blue line, but in a \textit{sense-in-motion} mode, where measurements are taken continuously during motion. The difference across these results highlights two key effects: (1) individual sensors can produce different responses under identical conditions, and (2) sampling in motion reduces signal magnitudes and shifts peak position due to shorter gas contact time and slow sensor dynamics~\cite{jin_design_2024}\cite{hoffman_rapid_2024}. Therefore, a direct comparison between estimated gas concentrations by DDPM and measured concentrations is nontrivial. While the absolute values of sensor readings vary across conditions,
the overall evolution remains similar. These observations
motivate us to consider the design of alternative features that
are robust to sensor variability and motion-induced artifacts. 
%While the absolute values of sensor readings vary across conditions, the overall evolution remains similar. These observations motivate us to consider the design of alternative features that are robust to sensor variability and motion-induced artifacts.
%The output voltage reading exhibits a nonlinear relationship with actual gas concentration, which is influenced by factors such as manufacturing variation, environmental conditions, and sensor aging~\cite{francis_gas_2022}. 

% Addressing this \textbf{model-to-measurement gap} involves either improving plume modeling fidelity or designing more robust gas features. Given the complexity and resource demands of the former, in this work, we propose to extract a gas feature that reduces dependency on precise plume models while effectively exploiting raw sensor readings and retaining information to ensure convergence of probabilistic GSL algorithms.

\subsection{Related Work}
Various gas features have been explored for GSL tasks, with the most commonly used ones being absolute gas concentration values and gas hit events. Several studies use concentration values directly, assuming a linear relationship between sensor output and actual concentration, modulated by a scaling factor 
%while accounting for the unknown gas release rate 
\cite{rahbar_design_2019,hutchinson2018information,zhu_novel_2020, jakkala_probabilistic_2022}. This model is only valid at low concentrations, where sensor nonlinearity is negligible~\cite{wang_metal_2010}. To address this point, \cite{wiedemann_model-based_2019} approximates the nonlinear response between the output voltage reading with actual gas concentration. However, this nonlinearity varies with gas type, manufacturing variability, environmental conditions (e.g., humidity, temperature, interfering gases), and sensor aging \cite{dennler_drift_2022}. As a result, each sensor must be calibrated with known concentrations of gas mixture in controlled conditions with a dedicated gas chamber to ensure accurate concentration approximation \cite{neumann_indoor_2019}\cite{turduev_experimental_2014}. Frequent calibration is required as the aforementioned environmental variations are uncontrollable, which makes it impractical in real-world settings. Gas hit events, binary indicators of gas presence or absence, which can be triggered by fixed~\cite{ojeda_information-driven_2021}\cite{wiedemann_gas_2022} or adaptive~\cite{li_odor_2011} thresholds, offer an alternative feature. 
%And some approaches use their rate as a feature ~\cite{vergassola2007infotaxis}\cite{bourne_decentralized_2020}. 
Compared to gas concentration values, gas hit events are more robust to sensor noise and drift. However, the binarization discards the gradient information from the measurements, and their accuracy depends heavily on the threshold selection.

\subsection{Problem Statement}
The objective of this work is to localize the 2D position of a single gas leak source in indoor environments with obstacles. We assume the leakage time is long enough for the plume to reach a steady state. The environmental layout and the inlet wind speed are assumed to be known. The input of the system are the gas concentration measurements gathered by a low-cost MOX sensor embedded on a mobile robot. At each iteration, the algorithm outputs a belief map indicating the likelihood of the source being in each part of the environment. The algorithm is terminated when the belief map converges or the maximum number of iterations is reached, and the final output is the estimated source location.
\subsection{Contribution}
In this work, we address the problem of localizing a single gas leak source with a ground robot equipped with a MOX gas sensor. To eliminate the need for tedious sensor calibration while preserving spatial gas intensity information, we introduce a novel rank-based gas feature that incorporates a probabilistic GSL algorithm. This method compares the relative ranking sequence of measured and estimated gas concentrations within their respective datasets, capturing structural similarity under the assumption of a nonlinear but monotonic relationship. By focusing on relative magnitude evolution rather than absolute values, this approach enables direct deployment on mobile robots endowed with uncalibrated MOX sensors.
More concretely, this paper presents the following contributions.
\begin{itemize}
    \item \textbf{MOX sensor simulation:} a faithful model that captures the nonlinear response and slow dynamics of a MOX sensor, integrated into the simulation pipeline.
    \item \textbf{Rank-based gas feature:} a novel gas feature that encodes the relative rank of each concentration measurement within the dataset. 
    \item \textbf{STE framework integration:} the integration of the proposed feature, along with three existing features, into a probabilistic Source Term Estimation (STE) framework~\cite{hutchinson_review_2017}, enabling comprehensive benchmarking of the gas feature selection.
    \item \textbf{Experimental validation:} a comprehensive evaluation of the resulting algorithmic pipeline through high-fidelity simulations and real-world experiments with different MOX sensors.
\end{itemize}
The outline of the paper is as follows: Section~\ref{sec:STE_framework} presents the proposed gas feature together with three alternative variants typically leveraged in the literature and their integration into a STE framework. 
%{\color{red}{Section~\ref{sec:mothodology} defines the various gas features}}.
Simulation and physical experiment results are detailed in Sections~\ref{sec:simulation} and \ref{sec:physical_exp}, respectively. Finally, Section~\ref{sec:conclusion} provides a discussion of the findings and future research directions.

\section{Methodology} \label{sec:STE_framework}
The STE algorithm is designed to estimate key parameters influencing the gas dispersion process, such as the source position, based on an inverse modeling approach \cite{hutchinson_review_2017}. 

\begin{figure}[t]
\vspace{0.2cm}
    \centering
    \includegraphics[width=1.0\linewidth]{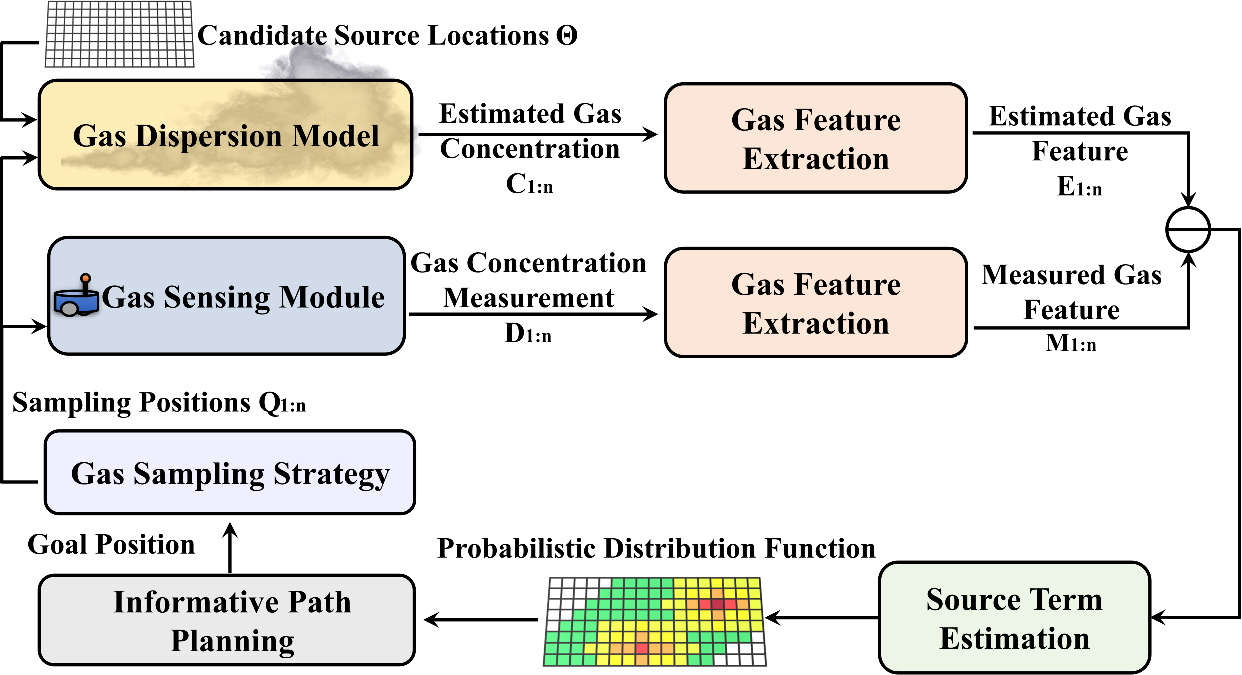}
    \caption{STE algorithm flowchart}
    \label{fig:STE_flowchart}
    \vspace{-0.6cm}
\end{figure}

%To formalize the estimation process, we define here key variables and notation used in the STE framework. 
\textit{Overview of the STE:} An overview of our STE algorithm with gas feature extraction is shown in Fig.~\ref{fig:STE_flowchart}. The environment is discretized into a grid, and each cell's probability of containing the source is iteratively evaluated. At each iteration, a steady-state gas dispersion model simulates plumes from every candidate cell and estimates the expected concentrations at the robot's sampling positions.
The alignment between gas features extracted from the \textit{Gas Sensing Module} and those from the \textit{Gas Dispersion Model} updates the probability of each cell containing the source. The resulting Probability Distribution Function (PDF) over source positions guides \textit{Informative Path Planning} algorithm, enabling more efficient data collection and accelerating estimation convergence. A new iteration begins once the robot reaches its goal and acquires a new batch of gas measurements. 

Let $\mathbf{\Theta}$ be a random variable representing the unknown source location. The environment is discretized into N grid cells, with candidate source locations represented by $\{\Theta_1, \Theta_2, ...\Theta_N\}$, with $\Theta_j$ denotes the center coordinates of cell $j$. For each iteration, let $D_{1:n}=\{d_1,...,d_n\}$ represent all the gas measurements gathered by the robots across sampling locations denoted by $Q_{1:n}=\{q_1,...,q_n\}$. For each possible source location $\Theta_j$, the gas dispersion model simulates the gas plume with the gas source located at $\Theta_j$ and derives the estimated gas concentrations represented by $C_{1:n}=\{c_1,...,c_n\}$ across the same set of sampling positions $Q_{1:n}$. The gas features extracted from the measured gas concentration $D_{1:n}$ are denoted by $M_{1:n}=\{m_1,...,m_n\}$, while those extracted from the estimated gas concentration $C_{1:n}$ are represented by $E_{1:n}=\{e_1,...,e_n\}$. Each component of the STE algorithm is detailed in the following sections.
\subsection{Gas Dispersion Model}
To incorporate the physical knowledge into the estimation process, a gas plume (dispersion) model is essential for predicting expected gas concentrations given a hypothesized source location $\Theta_j$.
To extend the applicability of STE to typical built environments, where closed-form model such as the Pseudo-Gaussian plume model~\cite{holmes_review_2006} can not be used, a DDPM was proposed in~\cite{jin_towards_2023} as a surrogate gas plume model. This approach leverages a Convolutional Neural Network (CNN) trained with gas dispersion maps under various environments with different obstacle configurations and source locations. The DDPM takes the obstacle configuration and hypothesized source location $\Theta_j$ as input and outputs the corresponding estimated gas concentration values {$C_{1:n}$. 
%Because of an emphasis on the generalization capability with different obstacle layouts, both training and testing datasets were collected under a fixed gas release rate and wind intensity. 
The same DDPM approach is leveraged in this work; for more information about its structure and training process, please refer to}~\cite{jin_towards_2023}.
%Training data is generated using wind simulations with different obstacle settings, conducted with the open-source computational fluid dynamic simulation software OpenFoam. The wind simulation results are then imported into the high-fidelity robotics simulator, Webots, to generate gas distribution maps under different obstacle configurations and random source locations. 
\subsection{Gas Sensing Module in Simulation}
 \begin{figure}[b]
\vspace{-0.2cm}
    \centering
    \includegraphics[width=\linewidth]{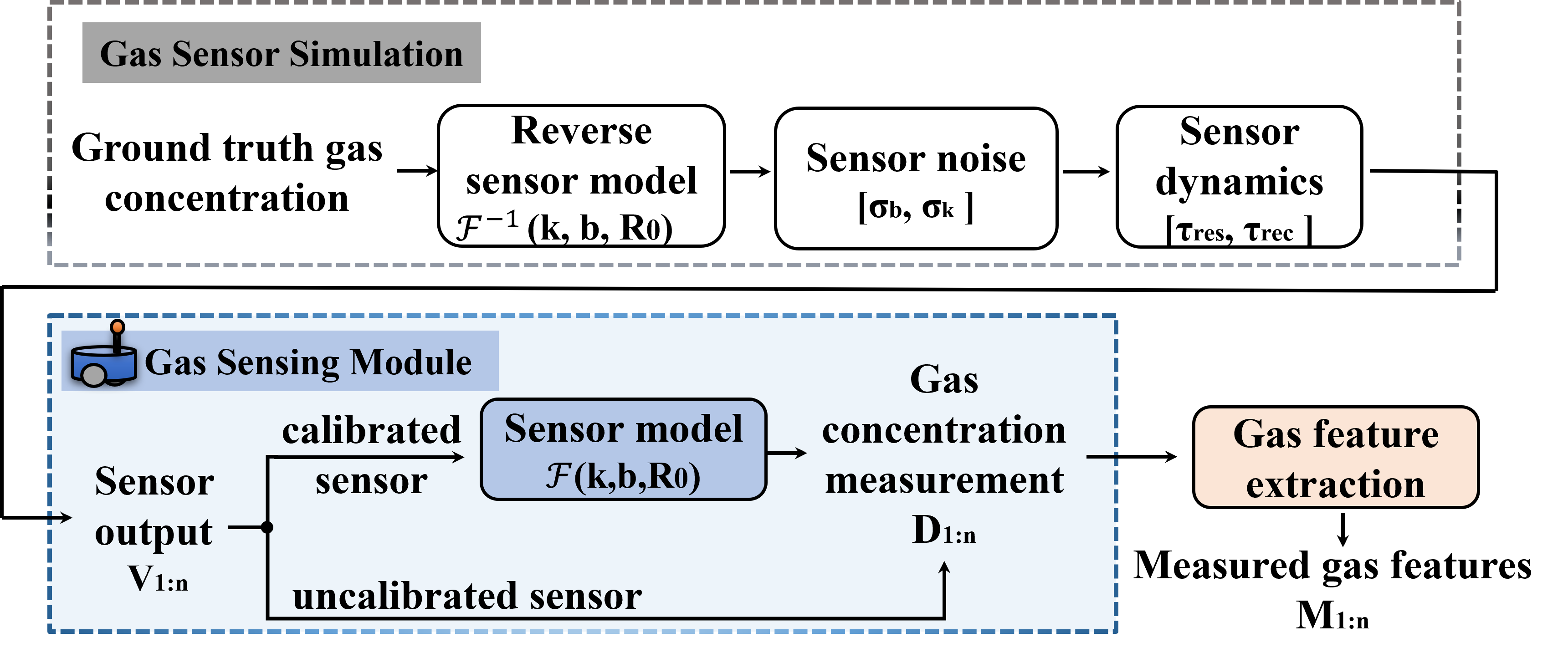}
    \caption{Data flow for the gas sensor simulation and gas sensing module}
    \label{fig:MOX_sensor}
\end{figure}
The robot samples the gas concentration at 10 Hz with an embedded MOX sensor. To have a faithful representation of the physical MOX sensor, three key characteristics are incorporated into the high-fidelity simulator. The simulated sensor model and the sensor data processing flowchart is shown in Fig.~\ref{fig:MOX_sensor}. 
\subsubsection{Nominal sensor model}
The MOX sensor comprises a heated metal oxide surface whose electrical resistance ($R_s$) changes in response to the interaction with oxidizing gases. A simple voltage divider circuit is used to measure $R_s$, incorporating a supply voltage $V_T$ and a load resistor $R_L$. The sensor output $V_{out}$ is the voltage measured across $R_L$, from which $R_s$ can be computed using the voltage divider equation.
% \begin{equation}
%     R_s = R_L * \frac{V_T - V_{out}}{V_{out}}
% \end{equation} 
According to the sensor datasheet, the logarithm of the ratio between $R_s$ and its baseline resistance in clean air $R_0$ exhibits a linear relationship with the logarithm of the gas concentration $g_{con}$. 
\begin{equation}
log(g_{con}) = k*log(\frac{R_s}{R_0}) +b
\end{equation}
The parameters $k$ and $b$ define the nonlinear sensor response and depend on gas type and sensor sensitivity. Due to manufacturing variations, each sensor may have different $R_0$ and sensitivity values. Additionally, $k,b,R_0$ are further influenced by environmental factors such as humidity and temperature. Combining these elements, the actual gas concentration $g_{con}$ can be approximated from the sensor output $V_{out}$ using:
\begin{equation}
g_{con} = \mathscr{F}(V_{out}, [k, b, R_0]) = 10^b * (\frac{R_LV_T-R_LV_{out}}{R_0V_{out}})^k
\label{eqa: sensor_calibration}
\end{equation}
The sensor model is nonlinear, and the parameters $k$, $b$ and $R_0$ must be obtained through calibration. Let $v_i$ and $d_i$ represent sensor output, which is voltage in our case,  and gas concentration measurement at the $i$th sampling position, respectively. To calculate $d_i$: 
\begin{itemize}
    \item For calibrated sensors, $d_i=\mathscr{F}(v_i, [k,b,R_0])$, where $\mathscr{F}(\circ)$ is the calibration function in Eq.~\ref{eqa: sensor_calibration}. In this case, gas concentration measurement corresponds to the approximate gas concentration.
    \item For an uncalibrated sensor, no transformation is applied, and the gas concentration measurement is obtained by directly taking the value of the sensor output $d_i=v_i$.
\end{itemize}
\subsubsection{Sensor noise}
The additive sensor noise $\sigma_s$ is modeled as the sum of two components: background noise $\sigma_b$ which has a fixed value, and gas intermittency noise $\sigma_k$, which is simulated as a proportional variation of the sensor output.
\begin{equation}
    \sigma_s = \sigma_b + \sigma_k*|v_i|
\end{equation}
with $\sigma_b$ and $\sigma_k$ are obtained via estimation from physical sensor outputs.
\subsubsection{Sensor dynamics}
MOX sensors exhibit slow dynamics, requiring a certain amount of time to transition from a baseline reading to a stabilized value when exposed to gas, and vice versa when exposed to clean air. Response and recovery phases of a MOX sensor are modeled as first-order systems, as described in \cite{monroy_overcoming_2012}\cite{monroy_gaden_2017}. The parameters of the simulated MOX sensor are calibrated to match the actual deployed sensor (MiCS-5521), with a response time $\tau_{res}$ and recovery time $\tau_{rec}$ of 2.04 s and 4.57 s, respectively.
% \begin{equation}
% R(s)= \begin{cases} \frac{K}{\tau_{res} s+1} & , \text{for response phase} \\ \frac{K}{\tau_{rec} s+1} & , \text{for recovery phase}\end{cases}
% \end{equation}
%The slow dynamics of the MOX sensor have minimal adverse effect when the stop-sense-go strategy is employed. However, in the sense-in-motion approach, they cause a spatial drift in sensor response relative to the original position of the gas stimulus~\cite{jin_sense_2024}.  
\subsection{Gas Features}
Gas features are used to evaluate the correlation between measured and estimated gas concentration. In the following, we present a novel gas feature based on concentration ranking, together with three existing ones typically adopted in previous studies.
\subsubsection{Gas measurement value}
The measured and estimated gas concentration values are directly adopted as gas features, which are defined as $m_i = d_i$ and $e_i = c_i$.
% \begin{equation}
% \begin{aligned}
% m_i &= d_i\\
% e_i &= c_i
% \end{aligned}
% \end{equation}
\subsubsection{Gas measurement hit}
A gas measurement hit is defined as a binary value determined by a threshold. It is formulated as $m_i = 1$ if $d_i>d_{thres}$ and $m_i=0$ otherwise,
% \begin{equation}
%     m_i =
%   \begin{cases}
%     1       & \quad \text{if } d_i>d_{thres},\\
%     0  & \quad \text{if } d_i < d_{thres},
%   \end{cases}
% \end{equation}
where $d_{thres}$ is the threshold for measured gas hits. The same formulation is applied to the estimated gas hits using a corresponding threshold $c_{thres}$. 

This threshold can be either fixed or adaptive. %Different values of $p_{thres}$ are tested, and when $p_{thres}=1000$, the algorithm performs the best in the default setting. 
We refer to the feature as \textit{fixed gas hit}, when a manually selected threshold is applied and remains constant throughout the experiments. %However, the thresholds need to be selected a priori, which is not suitable to unknown gas release rates and sensor model.
In the case of an adaptive threshold, which we refer to as \textit{adaptive gas hit}, we adopt the formulation proposed in~\cite{li_odor_2011}. The threshold $\bar{d}_{thres,i}$ is dynamically updated using a moving window average:
\begin{equation}
    \bar{d}_{thres,i} =
  \begin{cases}
    \lambda\bar{d}_{thres,i-1} + (1-\lambda)d_i       & \quad \text{if } i \geq 2,\\
    d_1  & \quad \text{if } i=1,
  \label{eqa: adaptive_gas_hit}
  \end{cases}
\end{equation}
The parameter $\lambda \in [0, 1)$ controls the smoothing factor. This feature enables detecting the gas hits at the rising edges of the smoothed gas concentration measurements. The same formulation is applied to the estimated gas hits, with $\bar{c}_{thres,i}$ computed from $C_{1:n}$. We choose $\lambda$ as 0.7 based on an exhaustive search, selecting the value that yields the best performance.
\subsubsection{Gas measurement rank} We propose using the relative measurement rank as the gas feature to mitigate the dependency on accurate gas concentration measurements (as assumed by the \textit{gas measurement value} feature) and avoid the loss of relative magnitude variation (as resulting from the \textit{gas measurement hit} feature). To this purpose, we compare the relative rankings of measured and estimated gas concentrations across all sampling points. Specifically, for a gas concentration measurement $d_i$ and its corresponding estimate $c_i$ at the same location, we compute their ranks within the respective datasets $D_{1:n}$ and $C_{1:n}$. This ranking-based approach captures the evolution of gas concentration patterns and assesses the alignment between measured and estimated values without relying on their absolute magnitudes. Because of this characteristic, it does not assume any specific functional relationship between measured and estimated gas concentration. It remain valid as long as a positive monotonic relationship is maintained, which means $D_{1:n}$ is considered to be aligned with $C_{1:n}$ as long as a higher measured gas concentration corresponds to a higher estimated gas concentration. To formalize this, we use the Empirical Distribution Function (EDF), which assigns a normalized rank to each measurement within the dataset without specific assumptions about the underlying data distribution. The EDF-based gas feature is defined as: 
\begin{equation}
    m_i= \frac{1}{n}\sum_{j=1}^n\textbf{1}_{d_j\leq d_i}
\end{equation}
where $\textbf{1}_A$ is the indicator of event A, which returns 1 if event $A$ is true and 0 otherwise. Thus, $m_i$ represents the proportion of measurements in $D_{1:n}$ that are less than or equal to $d_i$, yielding a value in the range $[0,1]$. This formulation transforms gas measurements into scale-invariant, rank-normalized features.
The same procedure is applied to the estimated gas concentrations $C_{1:n}$ to compute the EDF-based feature $e_i$.  Unlike the other three features, the EDF values $m_i$ and $e_i$ are dynamically updated at each iteration as new measurements are collected over successive iterations. Regarding the computational complexity, we do not need to resort the entire dataset at every iteration. Suppose $L$ measurements have been collected up to the previous iteration and their values are maintained in sorted order, and suppose $P$ new measurements are acquired in the current iteration. We first sort the $P$ new measurements, and then merge the two ranked lists. Therefore, the per-iteration computational complexity is $O(L + PlogP)$ (with $L+P=n$). In practice, when 
$L\gg P$, the update is dominated by the linear-time merge step.

% \textcolor{red}{Compared to the estimated gas concentration values, gas hit event and gas measurement rank features are more prone to identifying potential source positions with a perfect match when the observations are still sparse. This can cause the system to converge prematurely before sufficient data has been collected. To mitigate this issue, a modified likelihood function is introduced:}
% \begin{equation}
%     p(D_{1:n}|\mathbf{\Theta})\propto exp \bigg ( - \frac{1}{2}\sum_{i=1}^n \frac{n}{n+1}\frac{(m_i-s_i(\mathbf{\Theta}))^2}{\sigma_M^2+\sigma_S^2} \bigg )
%     \label{eqa:likelihood}
% \end{equation}
% Compared to Eq.~\ref{eqa: old_likelihood}, this formulation includes an additional term that accounts for the number of available observations, reducing the chance of premature convergence in the early stage of estimation.

\subsection{Estimation Process} 
For each candidate source location $\Theta_j$, the likelihood $p(D_{1:n}|\Theta_j)$ is calculated through the similarity across each $m_i$ and $e_i$ at the \textit{i}th sampling point. Here, $\sigma_M$ and $\sigma_E$ represent the standard deviations of the measurement and model estimation errors, respectively.
% \begin{equation}
%     p(D_{1:k}|\mathbf{\Theta})\propto exp \bigg ( - \frac{1}{2}\sum_{k=0}^N \frac{(d_k-c_k(\mathbf{\Theta}))^2}{\sigma_M^2+\sigma_E^2} \bigg )
%     \label{eqa:likelihood}
% \end{equation}
\begin{equation}
    p(D_{1:n}|\Theta_j)\propto exp \bigg ( - \frac{1}{2}\sum_{i=1}^n \frac{n}{n+1}\frac{(e_i(\Theta_j)-m_i)^2}{\sigma_E^2+\sigma_M^2} \bigg )
    \label{eqa:likelihood}
\end{equation}

Compared to the conventional likelihood function used in many existing works \cite{rahbar2019algorithm}\cite{hutchinson2018information}, we introduce in this work an additional $\frac{n}{n+1}$ term that accounts for the number of measurements. This adjustment mitigates the risk of premature convergence when measurements are sparse and scattered. Bayesian inference is used to update the posterior PDF $p\left(\mathbf{\Theta} \mid D_{1: n}\right)$, representing the probability distribution of the source position across $\mathbf{\Theta}$. 
%  \begin{equation}
%  p\left(\mathbf{\Theta} \mid {D}_{1: n}\right)=\frac{p\left(D_{1:n} \mid \mathbf{\Theta}\right) p\left(\mathbf{\Theta}\right)}{p(D_{1:n})}
% \end{equation}
We consider 
 %the evidence $p(D_{1:n})$ to be a normalization factor and 
 the prior $p(\mathbf{\Theta})$ to be a uniform distribution within the estimation area and assume a zero value inside the obstacles and boundary areas. %{\color{red} For each candidate source location $\Theta_j$, evaluating the likelihood in \eqref{eqa:likelihood} requires a single pass over the $n$ sampling points, and therefore costs $O(n)$. Updating the posterior over all $J = |\mathbf{\Theta}|$ candidates thus costs $O(Jn)$ per iteration (excluding the cost of generating the model estimates $e_i(\Theta_j)$).}
 %The process stops when either the maximum number of iterations is reached or the entropy of the PDF, which measures the estimation uncertainty, drops below a predetermined threshold.  

\subsection{Gas Sampling Strategy}
The \textit{stop-sense-go} strategy is commonly used in the literature due to gas plume intermittency and the slow sensor response. However, it reduces efficiency by prolonging the measurement process and discarding data collected during motion. In previous work \cite{jin_sense_2024}, the feasibility of employing a \textit{sense-in-motion} strategy is studied, where measurements $D_{1:n}$ are continuously taken on the way points $Q_{1:n}$ while the robot is moving. Despite increased noise in gas measurements, the increased measurement density compensates for reduced quality, resulting in a more efficient and robust sensing system. Therefore, we adopt the \textit{sense-in-motion} strategy in this work as well.
\subsection{Informative Path Planning}
The posterior PDF is used to calculate the information gain of the robot motion. Building on the sense-in-motion strategy, we adopt in this work a previously proposed informative path planning method called belief-clustering~\cite{jin_sense_2024}, which considers the information gained along the entire trajectory.
\subsection{Termination Condition}
The algorithm terminates when the entropy of the posterior PDF $p\left(\mathbf{\Theta} \mid D_{1: n}\right)$ drops below a predefined threshold and the set of candidate cells with high posterior probability is confined to a small spatial region, or when the maximum number of iterations is reached. Upon termination, the grid cell with the highest posterior probability is taken as the estimated source location. We set the maximum number of iterations to 30 in simulation and 40 in physical experiments. The higher limit in physical experiments accounts for the greater intermittency of real plumes, which makes gas measurements less reliable and slows convergence.
%%%%%%%%%%%%%%%%%%%%%%%%%%%%%%%%%%%%%%%%%%%%%%%%%%%%%%%%%%%%%%%%%%%%%%%%%%%%%%%%%%
% \section{{\color{blue}{MOX Sensor Modeling and Gas Feature Extraction}}} \label{sec:mothodology}
% This section introduces the MOX gas sensor model and defines the various gas features representing different levels of information extraction.
\section{Simulation Results} \label{sec:simulation}
\begin{figure}
\vspace{0.2cm}
    \centering
    \includegraphics[width=\linewidth]{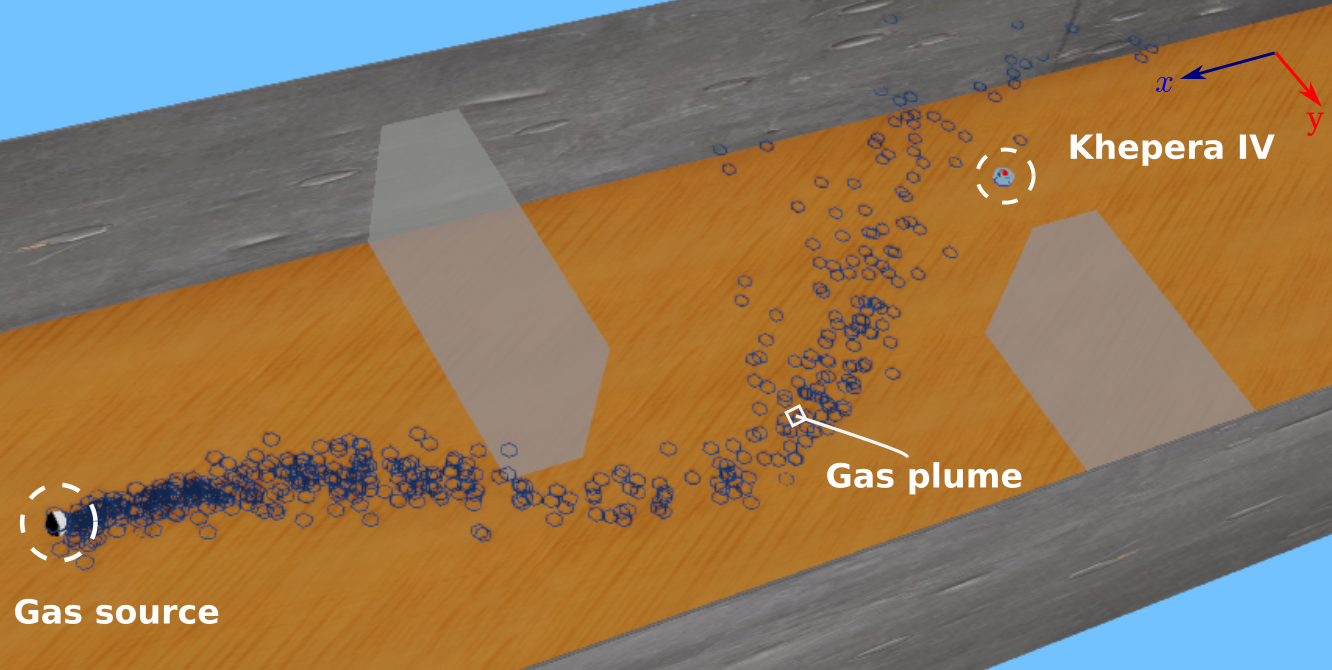}
    \caption{A screenshot from the Webots simulator displaying the environment with obstacles, gas plume and robot.}
    \label{fig:simulation_label}
    \vspace{-0.6cm}
\end{figure}

%The performance of the different gas features is evaluated considering both calibrated and uncalibrated MOX sensors. 
Simulations are conducted using the high-fidelity robotic simulator Webots\cite{Webots}, with a dedicated gas dispersion plugin \cite{webots_odor}. The testing environment includes two rectangular obstacles positioned centrally to create significant airflow disturbances, thereby influencing the gas dispersion dynamics, which is a typical environmental setup aligned with the work reported in \cite{jin_sense_2024}. The simulation setup is shown in Fig.~\ref{fig:simulation_label}. The performance of each feature is evaluated based on GSL accuracy and efficiency. GSL accuracy is determined by the source localization error, measured as the Euclidean distance between the estimated and true source positions. At each iteration, the robot updates its belief about the source position and selects the next goal position. To evaluate the GSL efficiency, the number of iterations required for the STE to converge is analyzed. Each experimental trial consists of ten runs with randomized source locations and robot starting positions. Out of brevity, we will use the terms ``\textit{gas value}", ``\textit{gas rank}", ``\textit{fixed gas hit}" and ``\textit{adaptive gas hit}" to refer to the gas features in the following text.
The performance of different gas features is evaluated using raw sensor voltage outputs from uncalibrated MOX sensors. For comparison, we also report results obtained with calibrated sensors. In the uncalibrated setting, we consider two different samples of MOX sensors, referred to as \textit{Sensor I} and \textit{Sensor II}, which differ in their base resistances $R_0$ and thus capture different levels of nonlinearity in the relationship between sensor output and actual gas concentration. We simulate the output voltage reading of uncalibrated MOX sensors through reverse sensor modeling, as demonstrated in Fig.~\ref{fig:MOX_sensor}. Specifically, two values of $R_0$ are considered: 100 $k\Omega$ and 1500 $k\Omega$, for \textit{Sensor I} and \textit{II} respectively, corresponding to the minimum and maximum factory variations specified in the sensor datasheet. The parameters $k=-1.58$ and $b=0.17$ are fixed based on datasheet calibration for ethanol detection under test conditions of 23 ± 5°C and 50 ± 10\% RH. In the calibrated setting, we use the same value of $[k,b,R_0]$ used for the reverse sensor modeling to calculate the calibrated gas concentration measurements. 

\begin{figure}[b]
\vspace{0.2cm}
    \centering
    \vspace{-0.4cm}
    \includegraphics[width=\linewidth]{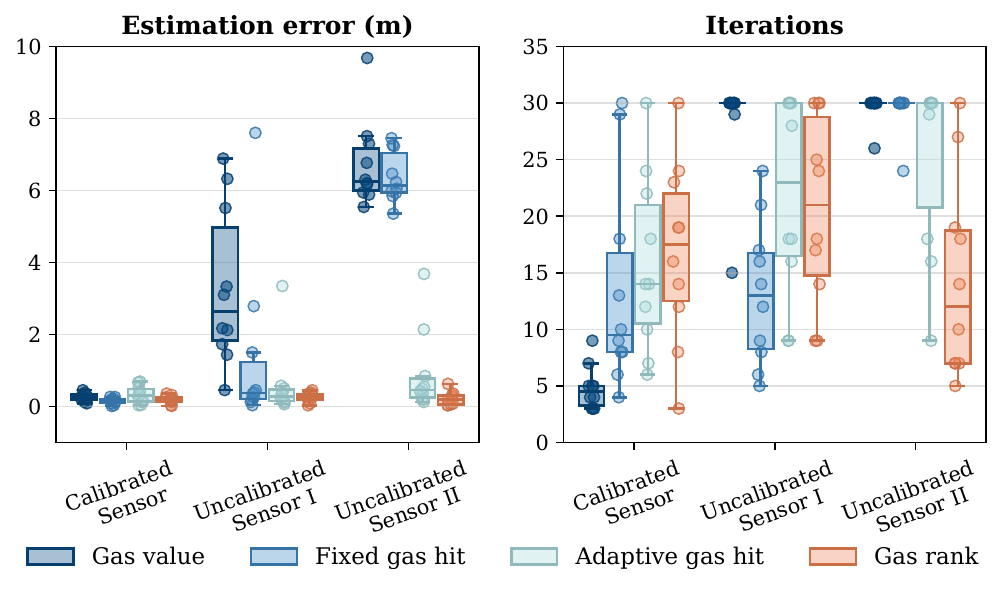}
    \caption{Performance comparison of the four different gas features with different gas sensors in simulation.}
    \label{fig:robustness_sensor_model}
    
\end{figure}
\begin{figure}
\vspace{0.2cm}
    \centering
    \includegraphics[width=\linewidth]{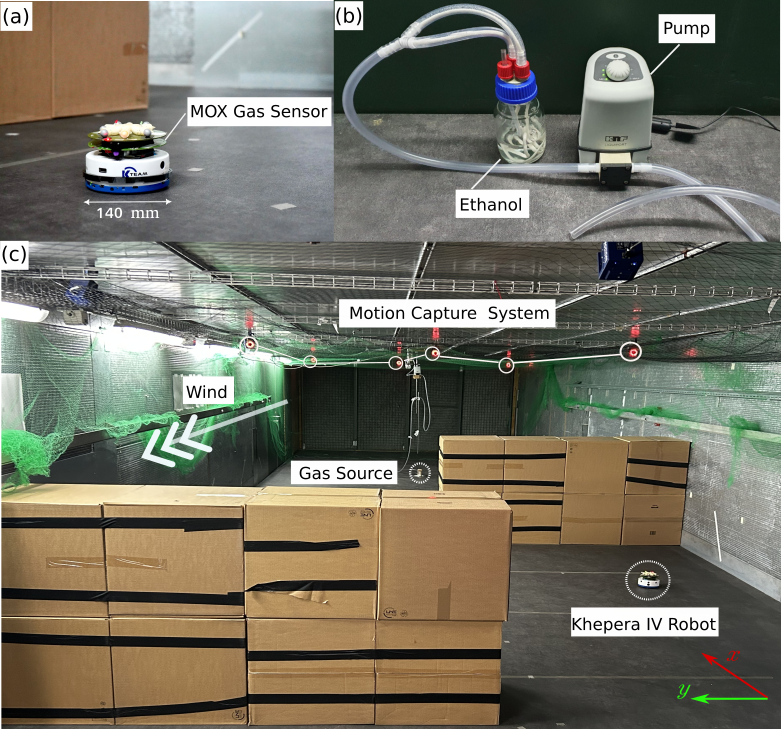}
    \caption{a) A Khepera IV robot equipped with the gas sensing module. b) Gas source. c) Experimental arena in the wind tunnel.}
    \label{fig:wind_tunnel}
    \vspace{-0.6cm}
\end{figure}

The results shown in Fig.~\ref{fig:robustness_sensor_model} indicate that all features achieve similar localization accuracy with the calibrated gas sensor. The \textit{gas value} feature converges faster by retaining most information from gas measurements and directly leverages magnitude differences in the likelihood function. When an approximated gas concentration closely matches the expected value near the gas source, rapid convergence occurs. 
In contrast, other features converge more slowly since the information is more abstract. However, the {\textit{gas value} feature lacks robustness with uncalibrated gas sensors, as gas concentration measurements, corresponding to raw sensor outputs in this case, no longer reflect actual gas concentrations. A similar limitation occurs with the \textit{fixed gas hit} feature, where the manually selected threshold fails to adapt to sensor models. In contrast, the \textit{adaptive gas hit} and \textit{gas rank} features remain consistent across different sensor models. They rely solely on a positive monotonic relationship between sensor outputs and gas concentrations, which preserves the rising edge and overall ranking of values despite nonlinearity. Here, we demonstrate the robustness of the proposed gas features under different sensor models by varying $R_0$. In principle, the same features can also generalize across different target gases, which would induce different sensor response curves (captured by different values of $k$ and $b$). This is because these features only require a positive monotonic relationship between sensor output and gas concentration, which is preserved for arbitrary $R_0$, $k$ and $b$.

Among these features, \textit{gas rank} outperforms \textit{adaptive gas hit} in localization accuracy and convergence efficiency. This is because sensor noise can lead to incorrect gas hit detections, and since \textit{adaptive gas hit} captures only local trends, it is more sensitive to these fluctuations. In contrast, \textit{gas rank} considers the entire data distribution, mitigating the impact of local noise and enhancing robustness.
%In summary, \textit{gas rank} demonstrates robustness across uncalibrated gas sensors. This robustness stems from its independence from absolute concentration values, while still preserving relative magnitude changes across sampling locations. 
Furthermore, compared to \textit{adaptive gas hit}, it requires no extra hyperparameter tuning ($\lambda$ in Eq.~\ref{eqa: adaptive_gas_hit}).
% \subsection{Summary}
% The \textit{gas concentration} feature performs best when the gas sensor is calibrated for the target gas and the release rate remains consistent with the training data. However, it lacks robustness when tested with unknown release rates or uncalibrated sensors, as it relies on matching the absolute values of expected and measured gas concentrations. The gas hit feature binarizes measurements, reducing dependency on precise gas concentration calculations. However, the \textit{fixed gas hit }feature does not generalize well across different testing conditions, requiring threshold adjustments for each scenario. In contrast, the \textit{adaptive gas hit} feature adapts dynamically based on the measurement set, making it more robust. However, it requires tuning of the smoothing parameter $\lambda$ to account for sensor noise and plume intermittency. The \textit{gas reading rank} feature eliminates the need for sensor measurements and model predictions to have the same absolute representation while preserving relative magnitude differences across sampling locations. It outperforms other methods in robustness and accuracy and does not require hyperparameter adjustments.

\section{Physical Experiments Results} \label{sec:physical_exp}
This section evaluates the \textit{gas value} (baseline) and \textit{gas rank} features employed with the STE algorithm for physical GSL experiments.
\subsection{Experimental Setup}
The experiments were carried out within our wind tunnel facility, ensuring consistent testing conditions across multiple runs. The wind tunnel has a volume of $18 \times 4 \times 1.9$ m and is equipped with a Motion Capture System (MCS) for accurate robot localization. The wind speed was maintained at 0.75 m/s throughout all experiments. The gas dispersion was generated using an electric pump vaporizing ethanol. 
%The pump draws air through an inlet connected to a bottle of ethanol via a tube and expels it through the outlet. The flow rate, adjustable via a continuous control knob, ranges from 0.2 to 1.3 L/min. 
We configured the wind tunnel with the same obstacle layout as in simulation, as shown in Fig.~\ref{fig:wind_tunnel}. A Khepera IV robot, equipped with a gas sensing module featuring a MiCS-5521 MOX sensor sampled at 10 Hz, was used as the mobile robotic platform. The robot's maximum speed was set to 0.27 m/s. The STE algorithm was executed on a laptop with an Intel Core i7-12700H processor. After each iteration, a new goal position was sent to the robot. The robot navigated to the goal point, collected gas samples continuously during motion, and transmitted them back to the laptop. The robot communicated with the laptop over high-speed WiFi. 

Given the effort required to conduct physical experiments for GSL tasks, we choose the \textit{gas value} feature as the baseline, as it is the canonical choice in most state-of-the-art STE methods} \cite{shi_autonomous_2025}\cite{jin_sense_2024}. To thoroughly evaluate the robustness of the \textit{gas rank} feature, we use two uncalibrated gas sensor boards (A and B) with distinct characteristics, as shown in Fig.~\ref{fig:sensor_scan}, 
%two gas release rates (default and high release rates), 
and varying source locations. 
%The default and high release rates correspond to pump speeds of 0.5 L/min and 1.3 L/min, respectively. Raw sensor outputs were used directly as gas concentration measurements. 
For each gas feature, fifteen experiments were conducted, with the gas source fixed along the $x$-axis, the $y$-axis varied across \{1, 2, 3\} meters to ensure broad spatial coverage.

\subsection{Results and Discussion}
% \subsubsection{Evaluation of the baseline feature} We evaluate the baseline feature, \textit{gas value}, using both calibrated and uncalibrated gas sensor modules under the default gas release rate. Each experimental trial includes five runs with a fixed source location and randomized robot starting positions. The results are presented in Tab.~\ref{table:baseline_exp}.

% \begin{figure}
% \vspace{0.2cm}
%     \centering
%     \includegraphics[width=0.5\linewidth]{img/Experiments_results/baseline_exp_calibrate_Vs_uncalibrate.eps}
%     \caption{Baseline experiments with gas concentration feature}
%     \label{fig:baseline_results}
%     \vspace{-0.6cm}
% \end{figure}
% \begin{table}[h!]
% %\vspace{0.2cm}
% \begin{center}
% \begin{tabular}{|c || c c|} 
%  \hline
%  Sensor calibration & Estimation error & Iterations \\ [0.5ex] 
%  \hline\hline
%  Calibrated & 0.43 [$\pm$0.41] & 3 [$\pm$1.8] \\ 
%  \hline
%  Uncalibrated & 5.24 [$\pm$2.29] & 21 [$\pm$2.86]  \\
%  \hline
% \end{tabular}
% \caption{Baseline experiments with \textit{gas value} as the gas feature.}
% \label{table:baseline_exp}
% \end{center}
% \vspace{-0.8cm}
% \end{table}
% From the result, it is clear that when sensors are properly calibrated to approximate actual gas concentrations, the \textit{gas value} feature achieves satisfactory localization performance, consistent with our previous results in~\cite{jin_sense_2024}. However, without calibration, \textit{gas value} fails to localize the gas source in all runs, confirming the simulation results.
% \subsubsection{Evaluation of the proposed feature}
\begin{figure}[t]
\vspace{0.3cm}
    \centering
    \includegraphics[width=\linewidth]{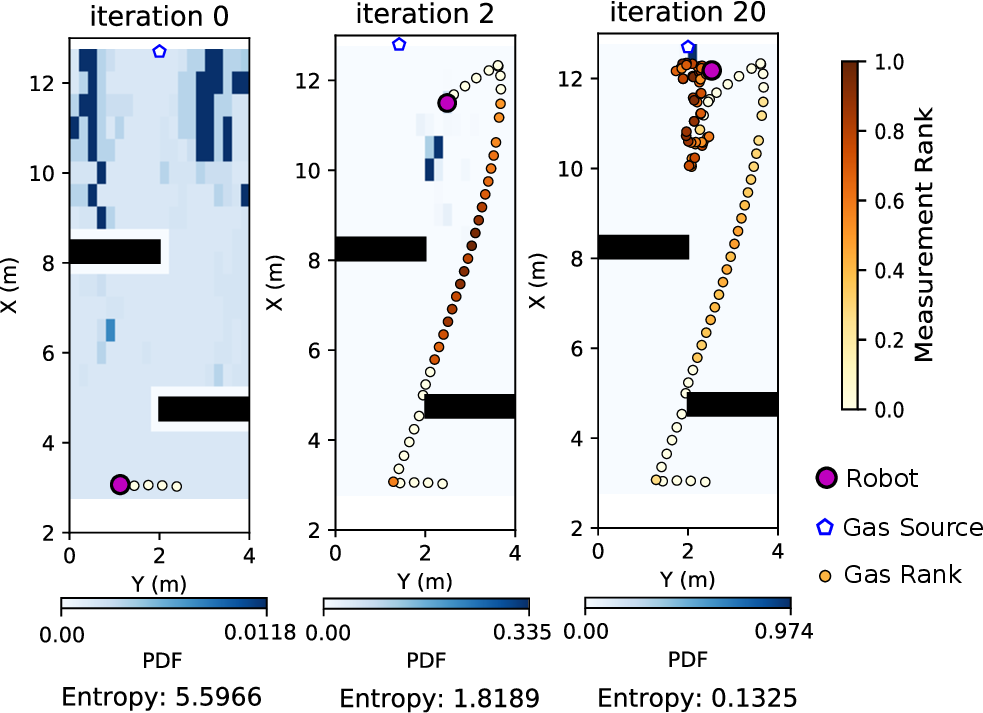}
    \caption{Illustration of the estimation process of one experimental run.}
    \label{fig:estimation_illustration}
\vspace{-0.4cm}
\end{figure}

An illustration of the estimation process of one experimental run using the \textit{gas rank} feature is shown in Fig.~\ref{fig:estimation_illustration}. The background color map in each subplot represents the current PDF over potential source locations. The circles along the robot’s trajectory represent gas measurements collected during motion, with their color indicating the EDF value, which is the rank-based gas feature. Over each iteration, the robot incrementally augments its dataset by exploring the environment. As new measurements are gathered, the relative EDF values are updated accordingly. Initially, the source belief is broadly distributed across the environment (left), but as more measurements are collected, the distribution becomes increasingly concentrated (middle), ultimately converging near the true source location (right).

\begin{figure}[t]
  \vspace{0.2cm}  
    \centering  \includegraphics[width=\linewidth]{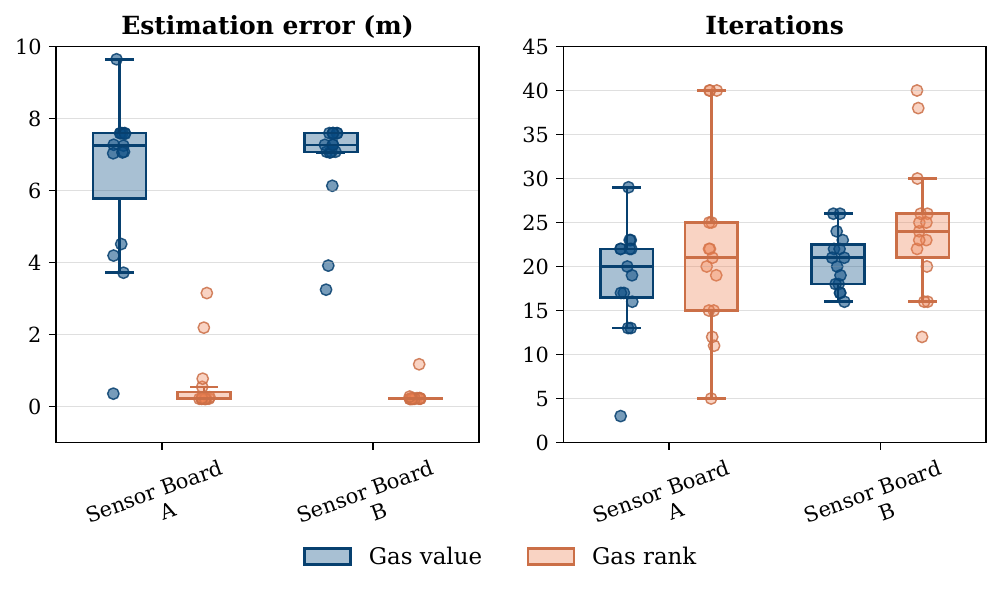}
    \caption{Physical experiments with two sensor boards.}
    \label{fig:robustness_reading_rank}
     \vspace{-0.4cm}  
\end{figure}

The results in Fig.~\ref{fig:robustness_reading_rank} show that the \textit{gas rank} feature successfully localizes the gas source with different gas sensors, without requiring sensor calibration and any parameter tuning. It performs consistently across sensor variations and source positions. In contrast, \textit{gas value}, which is typically used in the literature, fails to converge to the true source location with uncalibrated sensors. These findings align with the results observed in simulations. However, a higher localization error with some outliers is observed in real experiments compared to simulation. This discrepancy may be attributed to the increased intermittency due to turbulence of the gas plume and the accumulation of the gas in some region in the physical world, a phenomenon not faithfully reproduced in simulation. %At higher gas release rate, turbulence effects become more pronounced in the gas concentration, leading to noisier measurements in proximity to the source.

\section{Conclusion and Outlook} \label{sec:conclusion}
In this paper, we address the challenge of GSL using low-cost MOX gas sensors, which exhibit a nonlinear yet positive monotonic relationship between sensor outputs and actual gas concentrations. Due to their sensitivity to environmental parameters (e.g., temperature, humidity, other interfering gases), achieving and maintaining accurate sensor calibration is not practical. To overcome this limitation, we propose a novel \textit{gas rank} feature. It ranks measured gas concentration and compares the ranks to those of estimated concentrations derived from a gas plume model. This approach eliminates the need for sensor calibration to approximate the actual gas concentration values, as it only considers the relative shape of the data distribution across sampling locations. As a consequence, the proposed feature is robust to differences in magnitude and nonlinear mapping between sensor outputs and actual gas concentrations.

The proposed method is evaluated in both simulation and physical experiments and compared with three existing gas features: \textit{gas value}, \textit{fixed gas hit}, and \textit{adaptive gas hit}. Results show that both the \textit{adaptive gas hit} and \textit{gas rank} features demonstrate superior robustness to uncalibrated sensors. Between these two robust features, the \textit{gas rank} outperforms \textit{adaptive gas hit} in term of localization accuracy and convergence speed. Additionally, unlike other features, \textit{gas rank} does not require additional parameter tuning, such as the threshold for \textit{fixed gas hit} or the smoothing parameter ($\lambda$) for \textit{adaptive gas hit}. 

Given the fact that the only assumption to use this feature is the positive monotonic relationship between sensor outputs and actual concentration, this approach is adaptable to different sensor types and source estimation tasks. Moreover, as it depends solely on the ranking sequence of measurement values within respective datasets, it opens the door for multi-robot systems with heterogeneous gas sensors. In such settings, each robot can estimate source parameters based on its own gas concentration measurements and share probabilistic estimates with other robots without requiring cross-calibration of sensors.

Further improvements could enhance the convergence speed and reduce the outliers of the STE algorithm leveraging our novel \textit{gas rank} feature. Currently, only the rank of each measurement value is considered, but incorporating the slope (steepness) of the EDF function could provide additional information about how rapidly gas measurements changes spatially. This could lead to faster convergence and more robust localization of the source. %In parallel, we plan to extend our experimental setup to more complex and realistic environments, such as office-like settings, to better evaluate real-world applicability. This will include expanding the training dataset of our DDPM model to cover more diverse environmental layouts, a broader range of release rates, and varying wind conditions.
%%%%%%%%%%%%%%%%%%%%%%%%%%%%%%%%%%%%%%%%%%%%%%%%%%%%%%%%%%%%%%%%%%%%%%%%%%%%%%%%

%%%%%%%%%%%%%%%%%%%%%%%%%%%%%%%%%%%%%%%%%%%%%%%%%%%%%%%%%%%%%%%%%%%%%%%%%%%%%%%%

\bibliographystyle{IEEEtran}
\bibliography{IEEEexample}

\end{document}